\documentclass[11pt]{article}

\usepackage[final]{acl}

\usepackage{times}
\usepackage{latexsym}
\usepackage[most]{tcolorbox}
\usepackage[T1]{fontenc}

\usepackage[utf8]{inputenc}
\usepackage{subcaption}
\usepackage{eurosym}

\usepackage{microtype}
\usepackage{booktabs}
\usepackage{inconsolata}
\usepackage{graphicx}
\usepackage{fontawesome}
\usepackage{todonotes}

\title{\faBalanceScale ARCCS: An \textbf{A}utomated \textbf{R}egulatory \textbf{C}ompliance \textbf{C}hecking \textbf{S}ystem \\(Extended Version)}

\author{
Giorgos Filandrianos\textsuperscript{1,4}, 
José Menezes\textsuperscript{2}, 
Chrysoula Zerva\textsuperscript{2,4}, 
Alessandro Gianola\textsuperscript{3} \\
\textsuperscript{1}Instituto de Telecomunicações \\
\textsuperscript{2}Instituto de Telecomunicações/IST, Universidade de Lisboa \\
\textsuperscript{3}INESC-ID/IST, Universidade de Lisboa \\
\textsuperscript{4} AILS Lab;
    ECE, National Technical University of Athens \\ 
\href{mailto:geofila@islab.ntua.gr}{geofila@islab.ntua.gr}, \\
\href{mailto:\{jose.sacadura.menezes,chrysoula.zerva,alessandro.gianola\}@tecnico.ulisboa.pt}{\{jose.sacadura.menezes,chrysoula.zerva,alessandro.gianola\}@tecnico.ulisboa.pt}
}

\begin{document}
\maketitle
\begin{abstract}


Regulatory compliance checking---deciding whether a target document satisfies
the obligations of a regulation---requires interpreting dense legal text,
identifying which provisions apply, and grounding each decision in explicit
evidence. We present \textsc{ARCCS}, an end-to-end, automated, agentic, and
regulation-agnostic Legal NLP system for compliance checking. \textsc{ARCCS} decomposes
raw regulatory text into atomic, traceable requirements and evaluates a target
document against them using retrieved evidence, confidence scores, and
human-interpretable justifications. This design decouples compliance assessment
from any fixed regulatory template or predefined rule set, enabling the pipeline
to operate over regulations of varying size and structure. 
We evaluate ARCCS in two complementary settings. First, in a GDPR policy-document evaluation, LLM-based judges find its decisions and justifications legally and evidentially consistent in up to 96.67\% of the assessed cases. Second, on an EU public-procurement benchmark comprising more than 1,200 individual rule checks, the system attains 98.8\% accuracy in violation detection.
\textsc{ARCCS} is, to our knowledge, the first fully open-source system for end-to-end regulatory compliance checking and auditable report generation.

\end{abstract}

\section{Introduction}

Regulatory compliance checking---deciding whether a target document, such as a
vendor proposal or a terms-of-use agreement, satisfies the obligations imposed
by a body of regulation---is a high-stakes task that today rests almost
entirely on human experts. Doing it well requires three competencies that are
individually hard and jointly rare: \emph{interpreting} dense, cross-referential
legal text; \emph{identifying} which provisions are actually applicable to a
given document; and \emph{grounding} every decision in explicit textual
evidence so that it can be audited. This combination makes manual compliance
review slow and expensive, and difficult to scale to the document volumes of
modern procurement \cite{GianolaZ25}.

Legal NLP supplies strong components for these subtasks---domain-adapted
encoders and benchmarks for legal language understanding
\citep{chalkidis-etal-2020-legal, chalkidis-etal-2022-lexglue}, expert-annotated
contract review \citep{hendrycks2021cuad}, and broad legal-reasoning evaluation
of large language models (LLMs) \citep{guha2023legalbench}---and a parallel line
of work has shipped \emph{deployed} tools for individual legal artefacts:
unfair-clause detection in terms of service \citep{lippi2019claudette},
privacy-policy analysis \citep{harkous2018polisis}, and building-code compliance
\citep{hettiarachchi2025codeaccord}. These systems, however, are each tied to a
single artefact type or a single regulation, and do not offer an end-to-end path
from a raw, previously unseen regulation to an auditable, requirement-level
decision. Applying LLMs directly to close this gap is unsafe: models hallucinate
legal content at high rates \citep{dahl2024large}, and even retrieval-augmented
commercial legal tools remain unreliable \citep{magesh2024hallucination}. A
trustworthy system must therefore not only predict but also expose its evidence
and \emph{abstain} when support is insufficient---a behaviour studied as
answering only when warranted \citep{rajpurkar-etal-2018-know}, abstention in
LLMs \citep{wen-etal-2025-abstention}, and learning to defer to a human expert
\citep{madras2018predict, mozannar2020consistent}.

 We present \textsc{ARCCS}\footnote{\url{https://geofila.github.io/ARCCS/}}, an interactive system for end-to-end regulatory compliance checking. Given a regulation and a target document, \textsc{ARCCS} decomposes the regulation into atomic requirements, retrieves relevant evidence from the target document, and produces requirement-level compliance decisions with supporting excerpts and explanations. The system is designed to be regulation- and document-agnostic: it is not tied to a fixed checklist, a single legal domain, or a specific artefact type, but provides a reusable pipeline for turning complex regulatory text into traceable compliance assessments. 
 


This paper makes three contributions:
\begin{itemize}\setlength{\itemsep}{0pt}
    \item a general agentic compliance-checking algorithm that converts unseen
    regulations into atomic, traceable requirements and evaluates target
    documents using grounded evidence, abstention, and human deferral;

    \item an evaluation in two complementary settings: GDPR policy-document
    assessment with multiple LLM judges and an EU public-procurement benchmark
    of over 1{,}200 rule-level checks with deterministic ground truth; and

    \item to our knowledge, the first free and fully open-source implementation
    of an end-to-end regulatory compliance workflow, provided through an
    easy-to-use application that requires no programming or pipeline
    configuration and generates structured, auditable reports.
\end{itemize}
This paper is the extended version of a paper published at EMNLP 2026 (System Demonstrations) \cite{emnlp2026}.

\begin{figure*}[t]
    \centering
    \includegraphics[width=0.9\textwidth]{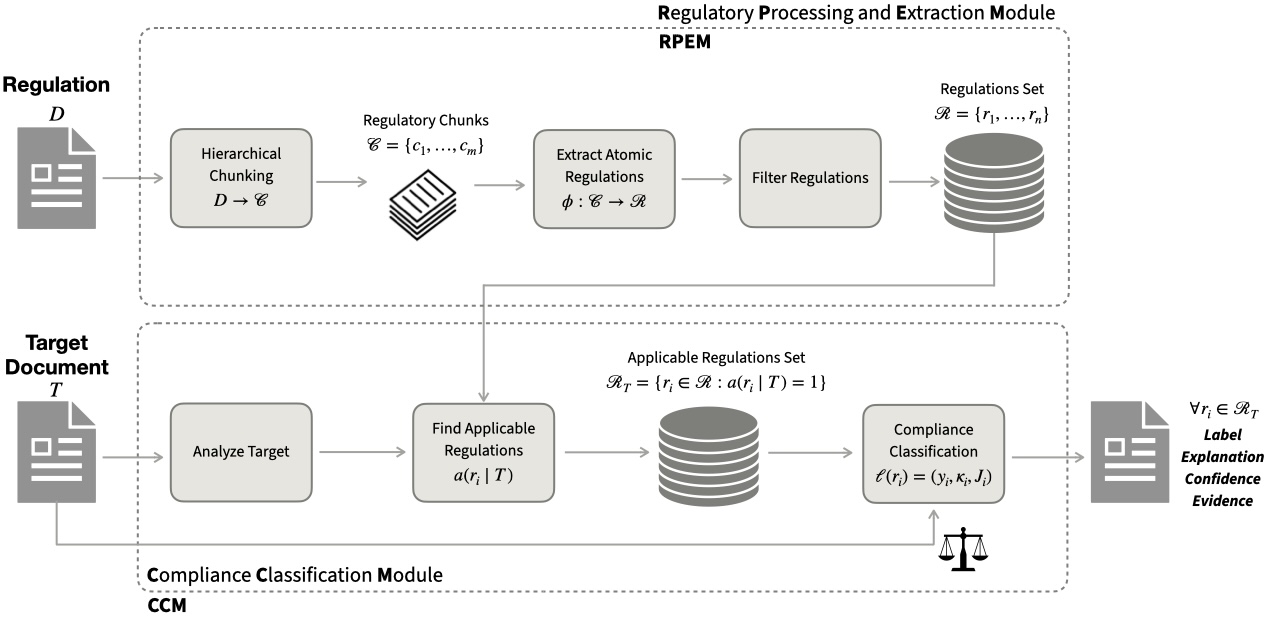}
    \caption{Overview of the proposed regulatory compliance assessment pipeline.}
    \label{fig:pipeline}
\end{figure*}

\section{System Architecture}
\label{sec:architecture}

This section describes what \textsc{ARCCS} computes and how. At a high level, \textsc{ARCCS} takes
two inputs---a regulatory document $D$ and a target document $T$ (a proposal or
terms-of-use agreement)---and returns a structured report that, for each
obligation in $D$, states whether $T$ satisfies it, with what confidence, and on
the basis of which textual evidence. As illustrated in
Figure~\ref{fig:pipeline}, computation proceeds in two stages. The
\emph{Regulatory Processing and Extraction Module} (RPEM, \S\ref{sec:rpem})
converts the regulation into a set of machine-checkable requirements; the
\emph{Compliance Classification Module} (CCM, \S\ref{sec:ccm}) checks the target
document against each requirement. We formalise the two modules below and then
give precise definitions of the four compliance labels (\S\ref{sec:labels}).

\subsection{Regulatory Processing and Extraction Module}
\label{sec:rpem}

\paragraph{Hierarchical segmentation.}
RPEM first segments the regulation $D$ into an ordered set of chunks
$\mathcal{C}=\{c_1,\dots,c_m\}$ that respects its legal structure (articles,
paragraphs, sub-paragraphs). Each chunk $c_j$ retains its structural metadata
(e.g.\ article identifier and position), so that local legal context is
preserved for fine-grained analysis, following structured legal-document
modelling \citep{chalkidis-etal-2020-legal}.

\paragraph{Atomic requirement extraction.}
RPEM then applies an extraction mapping
\begin{equation}
    \phi:\mathcal{C}\rightarrow\mathcal{R},
    \qquad
    \mathcal{R}=\{r_1,\dots,r_n\},
\end{equation}
that converts chunks into \emph{atomic} requirement objects. Each requirement is
represented as a tuple
\begin{equation}
    r_i=\bigl(f_i,\;\alpha_i,\;\sigma_i,\;M_i,\;\text{src}_i\bigr),
\end{equation}
where $f_i$ is the legal function of the provision (obligation, prohibition,
right, or condition), $\alpha_i$ the regulated actor, $\sigma_i$ the
jurisdictional and domain scope, $M_i=\{\mu_{i1},\mu_{i2},\dots\}$ the set of mandatory
conditions that must hold for the requirement to be satisfied, and $\text{src}_i$
an explicit pointer to the originating regulatory text. The mandatory-condition
set $M_i$ is what renders a requirement \emph{checkable}, and $\text{src}_i$ makes
every subsequent decision traceable to a specific provision 
---a deliberate response
to the evidence and auditability gaps behind legal hallucination
\citep{dahl2024large, magesh2024hallucination}.

\paragraph{Filtering and consolidation.}
Not all chunks yield enforceable constraints. RPEM removes provisions that are
purely definitional, descriptive, or contextual, and then deduplicates
requirements that are semantically equivalent but restated across
cross-references, merging them while preserving all source pointers. The result
is a compact, non-redundant requirement set $\mathcal{R}$ that serves as the
normative knowledge layer for the CCM.

\subsection{Compliance Classification Module}
\label{sec:ccm}

The CCM segments the target document into passages
$\mathcal{P}=\{p_1,\dots,p_k\}$ and evaluates each requirement
$r_i\in\mathcal{R}$ in three steps.

\paragraph{(1) Applicability.}
An applicability predicate
\begin{equation}
    a(r_i\mid T)\in\{0,1\}
\end{equation}
decides, from the scope $\sigma_i$ and the properties of $T$ (e.g.\ system type,
domain, deployment jurisdiction), whether $r_i$ governs $T$ at all. Requirements
with $a(r_i\mid T)=0$ are excluded from compliance reasoning, reducing analytical
noise and keeping the report focused on relevant obligations.

\paragraph{(2) Evidence retrieval.}
For each applicable requirement, the CCM scores the relevance of every passage
with a function
\begin{equation}
    \rho(r_i,p_j)\in[0,1],
\end{equation}
and selects the supporting evidence set
\begin{equation}
    E_i=\{\,p_j\in\mathcal{P}:\rho(r_i,p_j)\ge\tau\,\},
\end{equation}
where $\tau$ is a relevance threshold. The compliance decision for $r_i$ is then
conditioned on $E_i$ rather than on the model's parametric memory---a
retrieval-augmented design \citep{lewis2020rag} that grounds judgments in the
document text and is known to curb, though not eliminate, hallucinations \citep{magesh2024hallucination}.\looseness-1

\paragraph{(3) Labelling.}
Finally, a decision function
\begin{equation}
    \ell(r_i)=\bigl(y_i,\;\kappa_i,\;J_i\bigr)
\end{equation}
assigns a compliance label $y_i\in\mathcal{Y}$ (\S\ref{sec:labels}), a
confidence score $\kappa_i\in[0,1]$ derived from the consistency and
completeness of the evidence in $E_i$ with respect to the mandatory conditions
$M_i$, and a natural-language justification $J_i$ that cites the passages in
$E_i$ and the provision $\text{src}_i$.

\subsection{Compliance Labels}
\label{sec:labels}

\textsc{ARCCS} does not force a verdict on every requirement. Instead, the label space is
$\mathcal{Y}=\{\textsc{c},\textsc{nc},\textsc{ii},\textsc{hr}\}$, defined as
follows. Let $\theta_{\text{hi}}$ and $\theta_{\text{lo}}$ be confidence
thresholds with $\theta_{\text{lo}}<\theta_{\text{hi}}$.

\begin{description}
    \item[\textsc{Compliant} (\textsc{c}).] Every mandatory condition
    $\mu\in M_i$ is explicitly supported by the evidence $E_i$, no passage in
    $E_i$ contradicts $r_i$, and $\kappa_i\ge\theta_{\text{hi}}$. Formally,
    $\big(\forall\mu\in M_i:\ \text{sat}(\mu,E_i)\big)\ \wedge\
    \neg\,\text{conflict}(r_i,E_i)\ \wedge\ \kappa_i\ge\theta_{\text{hi}}$.
    \item[\textsc{Non-Compliant} (\textsc{nc}).] The evidence $E_i$ either
    directly contradicts a mandatory condition or omits a required condition
    while addressing its topic, with high confidence:
    $\big(\exists\mu\in M_i:\ \text{violate}(\mu,E_i)\big)\ \wedge\
    \kappa_i\ge\theta_{\text{hi}}$.
    \item[\textsc{Insufficient Information} (\textsc{ii}).] The evidence
    references the regulatory topic but does not determine satisfaction of the
    mandatory conditions; that is, some $\mu\in M_i$ is neither confirmed nor
    refuted by $E_i$, yielding low-to-moderate confidence
    ($\kappa_i<\theta_{\text{hi}}$). This is principled abstention: the system
    answers only when the evidence warrants it
    \citep{rajpurkar-etal-2018-know, wen-etal-2025-abstention}.
    \item[\textsc{Human Required} (\textsc{hr}).] A reliable automated decision
    cannot be made due to ambiguity, mutually contradictory evidence, missing
    external legal/technical context, or confidence below the deferral
    threshold ($\kappa_i<\theta_{\text{lo}}$). The case is thus routed to a human expert \citep{madras2018predict,mozannar2020consistent}.\looseness-1
\end{description}

The distinction between II and HR concerns the source of uncertainty:
\textsc{II} indicates missing documentary evidence, whereas
\textsc{HR} indicates that relevant evidence exists but requires
expert interpretation because it is ambiguous, conflicting, or
dependent on external context. An illustrative example follows.\looseness-1

\begin{center}
\fbox{%
\begin{minipage}{0.94\linewidth}
\small
\textsc{II:} The regulation requires an internal record of processing
activities, but the public terms provide no information about whether
such a record exists.

\smallskip
\textsc{HR:} The terms contain conflicting retention clauses, or the
decision depends on whether a legal exemption applies based on facts
outside the document.
\end{minipage}}
\end{center}

\paragraph{Compliance report.}
The output of \textsc{ARCCS} is a regulation-level report listing, for every applicable
requirement, the tuple $(y_i,\kappa_i,J_i,\text{src}_i)$. Since each entry
pairs a decision with its confidence, its supporting evidence, and the governing
provision, the report is auditable end-to-end and directly supports
human-in-the-loop review in real compliance workflows.\looseness-1

\section{Experimental Setup}
\label{sec:setup}

We evaluate \textsc{ARCCS} in two complementary settings. The GDPR setting
tests evidence-grounded assessment of policy documents without
deterministic gold labels, while the procurement setting tests
rule-level violation detection against independently implemented
deterministic checks.


\subsection{Setting 1: GDPR Policy-Document Compliance}
\label{sec:setup-gdpr}

\noindent\textbf{Inputs.}
We use the GDPR as regulatory input and evaluate the terms of use of three
widely used applications: WhatsApp, Netflix, and ChatGPT. The documents vary
substantially in length and specificity, allowing us to examine how available
textual evidence affects compliance labels. 

\noindent\textbf{Evaluation protocol.}
We report the predicted-label distribution over the 100 GDPR-derived requirements for each document. Since no deterministic gold standard is available, we additionally assess decision quality using an LLM-as-a-judge protocol \citep{gu2025survey,zheng2023judging}. For each sampled decision, the judge receives the regulation text, the extracted requirement representation, the relevant terms-of-use evidence, and the full \textsc{ARCCS} output, including the predicted label and explanation. We sample 30 decisions per document, yielding 90 judged instances, and use three independent evaluator models to assess consistency with the provided regulation and evidence as well as inter-evaluator agreement.



\subsection{Setting 2: Public-Procurement Event-Log Compliance}
\label{sec:setup-proc}

\noindent\textbf{Legal scope and benchmark.}
We evaluate public-procurement compliance under Directive~2014/24/EU
(CELEX:~32014L0024), focusing on twelve rules derived from Articles~4, 48, 56,
and 73 rather than the full Directive. The complete rule list and the operational mapping from legal provisions to executable checks are included with the released system.

\noindent\textbf{Data and ground truth.}
We construct a controlled benchmark of 100 procurement traces following the
Tenders Electronic Daily (TED) event-log schema.\footnote{\url{https://ted.europa.eu}}
For this, we take inspiration from previous work on the creation of event log data for public procurement processes \cite{NaiSG23,NaiSMGRG23,nai2025leveraging}. Because several relevant violations are sparse in naturally occurring data, the
benchmark includes 60 traces with at least one targeted violation and 40 fully
compliant controls. Expected labels are derived independently through
deterministic logic over each trace and its attributes, ensuring that
\textsc{ARCCS} is evaluated against externally defined legal-process ground
truth rather than model-generated judgments. This setting is therefore intended
as a controlled sanity check of rule-level compliance reasoning, not as evidence
of full real-world procurement coverage.

\noindent\textbf{Metrics.}
Each of the 100 traces is evaluated against all twelve rules, yielding
$1{,}200$ rule-level decisions. We report rule-level accuracy,
precision, recall, and $F_1$, treating true legal violations as the positive
class, together with case-level performance obtained by aggregating rule outputs
into a binary compliant/non-compliant trace label. We also report exact match,
defined as the proportion of cases for which all twelve rule decisions are
correct. The benchmark is evaluated with \texttt{gpt-5.4},
\texttt{gpt-5-mini}, and \texttt{gpt-5.2} under the same legal scope and
protocol.

\subsection{Models}
\label{sec:models}

\textsc{ARCCS} is model- and backend-agnostic: RPEM and CCM use structured prompts and typed outputs, without model-specific fine-tuning or local GPU requirements. In our experiments, we use OpenAI API models from the GPT-5 family, with \texttt{gpt-5.2} as the default backbone unless otherwise stated. All main runs use medium reasoning effort, and the confidence threshold for the \emph{Human Required} label is fixed at $0.60$. Exact model identifiers and model cards are reported in App.~\ref{app:model-cards}.


\section{Evaluation}
\label{sec:results}

\subsection{Policy-Document Compliance}
\label{sec:res-gdpr}
\textbf{Decisions are evidence-grounded.}
Table~\ref{tab:judge-acc} reports the accuracy of the three evaluator models on
the sampled GDPR set. The judges find \textsc{ARCCS} decisions consistent with the
provided regulation and evidence in 90.00--96.67\% of cases. Because the judge
conditions its verdict jointly on the regulation, the retrieved evidence, and
the \textsc{ARCCS} explanation, these scores indicate that the system produces
evidence-grounded rationales rather than unsupported label assignments.
Table~\ref{tab:judge-agree} reports inter-evaluator agreement: we observe
substantial overall agreement (Fleiss' $\kappa=0.69$, $p<0.001$), with
\texttt{gpt-5.2} and \texttt{gpt-5.2-pro} in perfect agreement ($\kappa=1.00$,
$p<0.001$), indicating that the favourable verdicts are stable rather than an
artefact of a single judge.

\begin{table}[t]
\centering
\small
\begin{tabular}{lc}
\toprule
\textbf{Evaluator (LLM-as-a-Judge)} & \textbf{Accuracy (\%)} \\
\midrule
\texttt{gpt-5.1}     & 96.67 \\
\texttt{gpt-5.2}     & 90.00 \\
\texttt{gpt-5.2-pro} & 90.00 \\
\bottomrule
\end{tabular}
\caption{Accuracy of model-based evaluators on the sampled set
(30 checks per terms-of-use doc.; 90 total).}
\label{tab:judge-acc}
\end{table}

\begin{table}[t]
\centering
\small
\begin{tabular}{lcc}
\toprule
\textbf{Agreement} & $\kappa$ & $p$-value \\
\midrule
Cohen's $\kappa$ (5.1 vs 5.2)      & 0.47 & 0.19 \\
Cohen's $\kappa$ (5.1 vs 5.2-pro)  & 0.47 & 0.19 \\
Cohen's $\kappa$ (5.2 vs 5.2-pro)  & 1.00 & $<0.001$ \\
Fleiss' $\kappa$ (3 raters)        & 0.69 & $<0.001$ \\
\bottomrule
\end{tabular}
\caption{Inter-evaluator agreement on the binary verdict (\textsc{yes}/\textsc{no})
for the sampled set.}
\label{tab:judge-agree}
\end{table}


 \noindent\textbf{Abstention is evidence-faithful, not evasive.} Across the GDPR-derived requirements, \emph{Insufficient Information} is the dominant outcome, as expected for public terms-of-use documents. Most GDPR obligations depend on internal governance artefacts, such as records of processing, DPIAs, and security measures, which are not normally disclosed in public-facing terms. A faithful compliance system should therefore abstain when the document does not provide enough evidence, rather than infer internal practices from silence. Importantly, abstention does not prevent the system from issuing determinate verdicts when the retrieved evidence supports them: \textsc{ARCCS} reserves \emph{Compliant}/\emph{Non-Compliant} labels for evidence-backed cases and routes borderline cases to \emph{Human Required}.\looseness-1

 \noindent\textbf{Surfacing confirmed conflicts.} Beyond missing information, \textsc{ARCCS} surfaces \emph{candidate} normative conflicts for human review---system-generated flags, not legal determinations. Some of these flags correspond to conflicts that have independently been confirmed by regulators. For example, \textsc{ARCCS} flags WhatsApp terms as problematic, and the Irish Data Protection Commission later found WhatsApp non-compliant and imposed a \euro{}225M fine under Arts.~5(1)(a) and 12--14 GDPR.\footnote{\href{https://www.dataprotection.ie/en/news-media/press-releases/data-protection-commission-announces-decision-whatsapp-inquiry}{Irish Data Protection Commission WhatsApp decision.}} Similarly, for the \emph{ChatGPT Terms of Use}, \textsc{ARCCS} flags GDPR~Article~79 as \emph{Non-Compliant}, because mandatory-arbitration and exclusive San Francisco forum clauses conflict with the Article~79 guarantee that EU/EEA data subjects may sue in their Member State of residence. These cases show that the system identifies concrete evidence-backed conflicts, not only gaps, while still leaving final legal interpretation to experts.

\subsection{Public-Procurement Logs}
\label{sec:res-proc}
\noindent\textbf{Rule-level performance.}
At rule level (Table~\ref{tab:rule}) all three models perform strongly and
remain tightly clustered: accuracy ranges from 98.5\% to 98.8\% and recall is
near-saturated (98.8--100.0\%), indicating robust detection of true violations.
\texttt{gpt-5.2} and \texttt{gpt-5-mini} jointly provide the strongest profile
(precision 85.9, recall 100.0, $F_1$ 92.4), outperforming \texttt{gpt-5.4}
chiefly through fewer false alarms. Crucially, the residual error is
false-positive-driven rather than miss-driven: since recall is near 100\%,
\textsc{ARCCS} follows the conservative error
profile that compliance screening demands.

\begin{table}[t]
\centering
\small
\begin{tabular}{lcccc}
\toprule
\textbf{Model} & \textbf{Acc.} & \textbf{Prec.} & \textbf{Rec.} & \textbf{$F_1$} \\
\midrule
\texttt{gpt-5.4}    & 98.5 & 83.2 & 98.8  & 90.3 \\
\texttt{gpt-5.2}    & 98.8 & 85.9 & 100.0 & 92.4 \\
\texttt{gpt-5-mini} & 98.8 & 85.9 & 100.0 & 92.4 \\
\bottomrule
\end{tabular}
\caption{Rule-level performance on the procurement benchmark
(100 traces, 1{,}200 rule decisions per model).}
\label{tab:rule}
\end{table}

\noindent\textbf{Case-level performance.}
Aggregated to a per-trace violation/no-violation decision
(Table~\ref{tab:case}), performance is strong: \texttt{gpt-5.2} and
\texttt{gpt-5-mini} reach 100.0\% on all case-level metrics, while
\texttt{gpt-5.4} attains 98.0\% accuracy and 98.4\% $F_1$, correctly flagging
all violating traces. As a first-stage screening mechanism that aims to separate
problematic from compliant cases \textsc{ARCCS} is therefore highly reliable.

\begin{table}[t]
\centering
\small
\begin{tabular}{lcccc}
\toprule
\textbf{Model} & \textbf{Acc.} & \textbf{Prec.} & \textbf{Rec.} & \textbf{$F_1$} \\
\midrule
\texttt{gpt-5.4}    & 98.0  & 96.8  & 100.0 & 98.4 \\
\texttt{gpt-5.2}    & 100.0 & 100.0 & 100.0 & 100.0 \\
\texttt{gpt-5-mini} & 100.0 & 100.0 & 100.0 & 100.0 \\
\bottomrule
\end{tabular}
\caption{Case-level (per-trace) performance on the procurement benchmark.}
\label{tab:case}
\end{table}

\noindent\textbf{Exact-match performance.}
Under the strictest criterion---a trace counts as correct only if all twelve
rule decisions are simultaneously correct---scores are necessarily lower
(Table~\ref{tab:exact}), since a single rule-level slip invalidates the trace.
\texttt{gpt-5-mini} is best (87.0\%) and also exhibits the lowest
\emph{Insufficient Information} rate (2.3\%, vs 7.3\% for \texttt{gpt-5.2} and
8.8\% for \texttt{gpt-5.4}), indicating slightly better fine-grained calibration
when all rules must hold at once. The gap between near-perfect case-level scores
and exact-match scores stems from isolated extra false alarms on individual
rules rather than broad misunderstanding of entire traces, and should be read as
a fine-grained calibration challenge rather than a failure of core violation detection.\looseness-1

\begin{table}[t]
\centering
\small
\begin{tabular}{lc}
\toprule
\textbf{Model} & \textbf{Exact Match (\%)} \\
\midrule
\texttt{gpt-5.4}    & 83.0 \\
\texttt{gpt-5.2}    & 86.0 \\
\texttt{gpt-5-mini} & 87.0 \\
\bottomrule
\end{tabular}
\caption{Document-level exact-match performance: the percentage of traces for
which all 12 rule decisions are simultaneously correct.}
\label{tab:exact}
\end{table}


\section{Demonstration}
\label{sec:demo}

\textsc{ARCCS} is delivered as a self-contained web application that runs locally and
requires no programming, model configuration, or pipeline orchestration from the
user: after supplying their own API key, the user starts the server and reaches
the interface in a browser, where the full \textsc{ARCCS} pipeline is driven
through a three-step wizard. \emph{Step~1}, the user uploads a regulatory
document or selects one of the preloaded regulations (e.g., the GDPR), which
skips extraction and reuses a cached, pre-extracted requirement set.
\emph{Step~2}, the user uploads the target policy or terms-of-use document via
drag-and-drop. \emph{Step~3}, the user launches the analysis and the system
checks the document against the extracted requirements. Throughout extraction
and checking, the interface streams real-time logs that expose pipeline progress
and intermediate stages, making the reasoning process transparent to
non-technical stakeholders.

The user inspects the outcome through a structured compliance report. A summary
panel reports the counts of the four labels
together with an aggregate compliance-rate indicator; below it, a results table
lists one row per requirement. Selecting any row opens a detail view exposing the
governing provision and its identifier, the assigned label, the confidence
score, the verbatim evidence quoted from the target document, and the
natural-language justification. This layout makes the outputs directly
interpretable: a non-expert can read a verdict, see the exact clause that
supports it, and immediately recognise the cases the system has deferred---those
labelled \emph{Human Required} (confidence below the $0.60$ threshold) or
\emph{Insufficient Information}---as the ones warranting manual review. Because
each decision is linked to its source provision and evidence, the report is
auditable clause by clause. The user can export the complete report as a
structured JSON file, and the system persists a browsable history of prior runs,
enabling iterative auditing and comparison across documents and over time. The
application itself runs locally: documents are never uploaded to a third-party web
service, and the only external call is to the language-model provider chosen by
the user, which can be a self-hosted open-weight model for fully air-gapped
deployments. A step-by-step visual walkthrough, worked example outputs, and
implementation details are provided in App.~\ref{app:demo},
\ref{app:examples}, and~\ref{app:impl}. Additionally, the released system includes the full rule definitions, implementation details, prompts, example inputs and outputs, and materials needed to reproduce the demo.

\section{Conclusion}
\label{sec:conclusion}

We presented \textsc{ARCCS}, an end-to-end interactive system for regulatory compliance checking that turns a regulation and a target document into an auditable compliance report. By decomposing the regulation into atomic, traceable requirements, \textsc{ARCCS} can analyse long legal and policy documents requirement by requirement, grounding each decision in retrieved evidence, confidence scores, and explanations rather than relying on a fixed checklist or hand-crafted rules. Our evaluation demonstrates strong performance in both policy-document compliance and deterministic public-procurement checking, showing that the same architecture supports evidence-grounded assessment and accurate rule-level violation detection. Together with the openly available, on-premise demo, this positions \textsc{ARCCS} as a transparent tool for human-in-the-loop regulatory review.

\section*{Limitations}
\label{sec:limitations}
Several limitations qualify our findings and outline directions for future work.

\noindent\textbf{Absence of a human gold standard in the GDPR setting.}
For the terms-of-use experiments no deterministic ground truth exists, so we
rely on an LLM-as-a-judge protocol. While inter-judge agreement is substantial
overall (Fleiss' $\kappa=0.69$), agreement between some evaluator pairs is only
moderate and not statistically significant (Cohen's $\kappa=0.47$, $p=0.19$).
Moreover, all evaluators
belong to the GPT-5 model family, and GPT-5.2 is also used as the default
\textsc{ARCCS} backbone. This limits the independence of the evaluation. A rigorous human evaluation by qualified legal
experts is needed to validate these results and is left to future work.

\noindent\textbf{Controlled data and limited real-world validation.}
Because violations are sparse in real procurement logs, the procurement
benchmark consists of controlled traces built on the TED schema with
deliberately balanced violations. This design enables reproducible,
deterministic evaluation, but performance on naturally occurring, noisier logs
may differ, and the externally defined ground truth reflects our operational
formalisation of the Directive rather than adjudicated legal outcomes.

\noindent\textbf{Component-level evaluation.}
We therefore do not include dedicated comparisons against simpler
direct-prompting or vanilla RAG configurations, nor systematic ablations of the
individual stages of the pipeline. In particular, the fidelity of the requirements
produced by RPEM is not independently evaluated. Errors introduced during requirement
extraction may propagate to downstream compliance decisions even when the
subsequent CCM reasoning is internally consistent.

\noindent\textbf{Limited scale and scope of evaluation.}
The judged GDPR sample comprises 90 instances, and the quantitative benchmark
covers 100 traces under a single Directive. Our study also spans only two
regulatory regimes (the GDPR and Directive~2014/24/EU) in English; generalisation
to other legal domains, jurisdictions, and languages remains to be demonstrated.

\noindent\textbf{Model coverage and configuration.}
Although \textsc{ARCCS} is designed to be model-agnostic, the reported experiments
use only the GPT-5 family through a commercial API. We have not yet quantified
performance with open-weight, fully on-premise backbones. Likewise, the decision
thresholds (the \emph{Human Required} confidence threshold of $0.60$ and the
retrieval threshold $\tau$) were fixed rather than systematically calibrated.
Finally, although violation recall is near-perfect to perfect in the
controlled procurement benchmark, this evaluation does not establish the prevalence
of the more consequential false-Compliant failure mode---that is, cases in which a
true violation is not surfaced as \emph{Non-Compliant}---in naturally occurring
regulatory settings.

\noindent\textbf{Scope of intended use.}
\textsc{ARCCS} is a decision-support tool, not a substitute for professional legal
judgment. Its outputs are intended to assist expert review through transparent,
evidence-linked reports, and consequential compliance determinations should
remain subject to human oversight.

\section*{Ethical Considerations}
\textsc{ARCCS} is a decision-support tool and is not a substitute for professional legal
judgment; it must not be used to make final, consequential compliance decisions
without expert oversight. Several risks follow directly from the task. (i)~\textbf{False
negatives}---a missed violation---could create a false sense of safety; we
therefore tune the system conservatively (recall at or near $1.0$ in our
procurement experiments) and surface confidence and evidence so users can
prioritise review. (ii)~\textbf{False positives and hallucinated legal
interpretations} could mislead non-expert users; to mitigate this, every output
is grounded in retrieved source text, every \textsc{nc} flag is presented as a
candidate conflict for human verification, and the \emph{Insufficient
Information} and \emph{Human Required} labels make abstention explicit rather than
forcing a verdict. (iii)~\textbf{Over-reliance and accountability}: the interface
presents verdicts as reviewable flags with citations, and we document that legal
responsibility remains with the human user. (iv)~\textbf{Privacy}: the
application runs locally and processes documents on the user's own machine; when a
hosted model API is used, only the user's chosen provider receives the text, and a
self-hosted open-weight model can be used for fully air-gapped operation. We make
the code and data public to support scrutiny and reproducibility, and we caution
that deployment in a new jurisdiction or legal domain requires validation by
qualified legal experts.

 \section*{Acknowledgments}
 This work was partially supported by the `OptiGov' project, with ref. n. 2024.07385.IACDC (DOI: 10.54499/2024.07385.IACDC), fully funded by the `Plano de Recuperação e Resiliência' (PRR) under the investment `RE-C05-i08 - Ciência Mais Digital' (measure `RE-C05-i08.m04'), framed within the financing agreement signed between the `Estrutura de Missão Recuperar Portugal' (EMRP) and Fundação para a Ciência e a Tecnologia, I.P. (FCT) as an intermediary beneficiary. A. Gianola was also partly supported by Portuguese national funds through Fundação para a Ciência e a Tecnologia, I.P. (FCT) under projects UID/50021/2025 (DOI: https://doi.org/10.54499/UID/50021/2025) and UID/PRR/50021/2025 (DOI: https://doi.org/10.54499/UID/PRR/50021/2025).
G. Filandrianos and C. Zerva were also partly supported within the framework of the Pharos AI Factory project, funded by the European High-Performance Computing Joint Undertaking (EuroHPC JU) under Grant Agreement No. 101234269 as part of the Horizon Europe and by the Greek Public Investments Program programme.

\bibliography{custom}

\clearpage
\appendix

\section{Demonstration Walkthrough}
\label{app:demo}
\begin{figure*}[t]
\centering
\includegraphics[width=\linewidth]{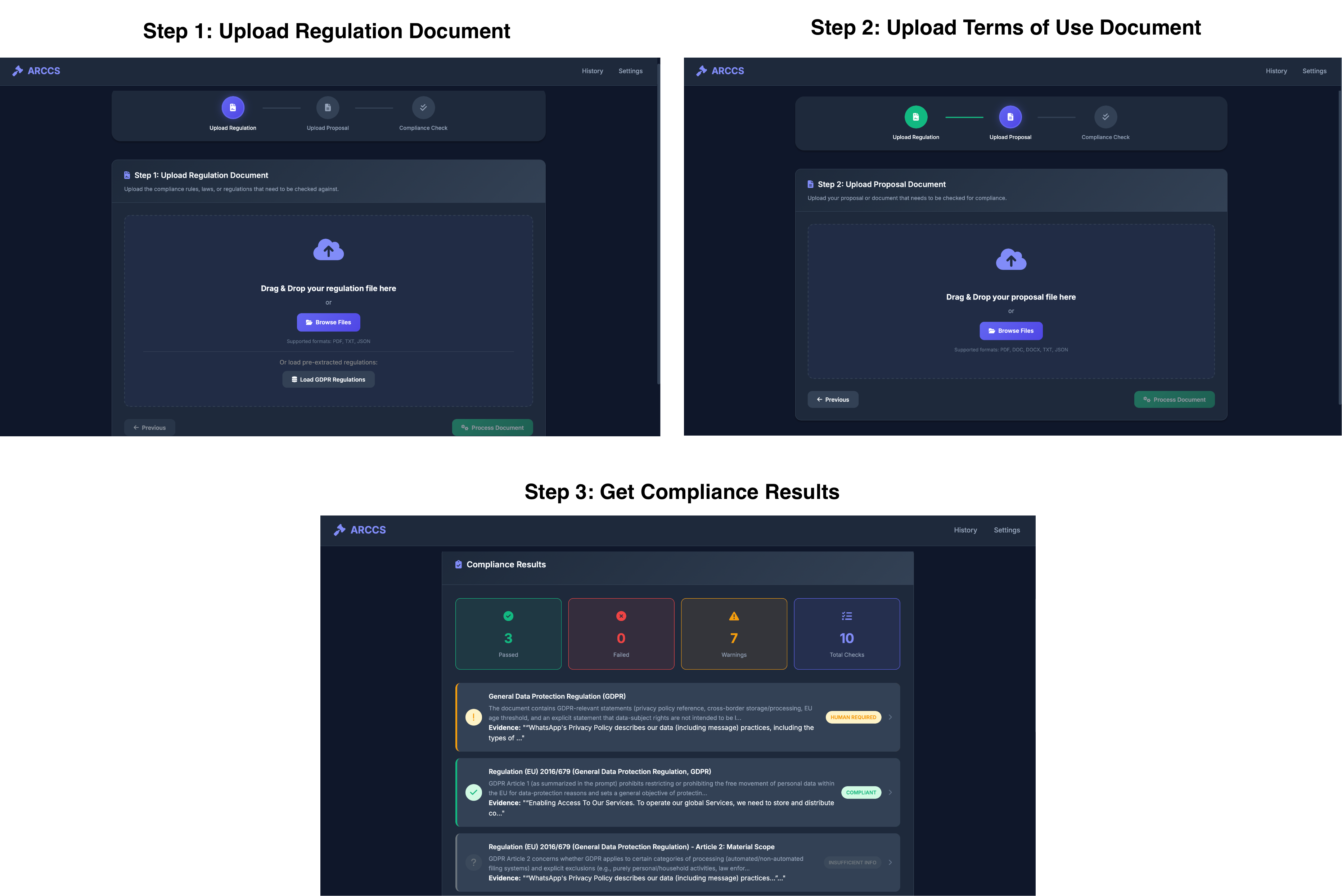}
\caption{The \textsc{ARCCS} web demonstration interface, designed for users without
programming experience. The workflow takes a regulation and a target document
as input and returns an auditable, evidence-linked compliance report.}
\label{fig:demo}
\end{figure*}

This appendix provides a visual walkthrough of the \textsc{ARCCS} web interface
(Figure~\ref{fig:demo}), which is designed for users without programming
experience. The interface abstracts away model configuration and pipeline
orchestration, exposing the end-to-end compliance workflow through three
sequential steps (Figure~\ref{fig:demo}).

\paragraph{Step 1 -- Upload the regulation.}
The user uploads a regulatory document or selects one of the preloaded
regulations, such as the GDPR. Internally, this input is passed to the RPEM,
which performs hierarchical segmentation and atomic requirement extraction.

\paragraph{Step 2 -- Upload the target document.}
The user uploads the policy or terms-of-use document to be assessed. This
document is segmented into passages that serve as the evidence pool for the
retrieval-augmented compliance reasoning performed by the CCM.

\paragraph{Step 3 -- Inspect the compliance report.}
After execution, the interface renders a structured compliance report. A summary
panel reports the counts of the four
compliance labels, followed by a per-requirement breakdown in which each entry
shows the governing provision, the assigned label, the confidence score, the
retrieved evidence, and a natural-language justification. During execution, the
interface additionally streams real-time logs that expose pipeline progress,
and it preserves a history of prior runs to support iterative auditing and
cross-document comparison.

\section{Qualitative Example}
\label{app:examples}
This appendix gives representative \textsc{ARCCS} outputs for all labels, using GDPR checks against WhatsApp terms of use.


\begin{tcolorbox}[
  title=Example 1 -- Compliant (GDPR Article 2),
  colback=green!3, colframe=green!55!black,
  fonttitle=\bfseries, breakable, enhanced, sharp corners, boxrule=0.6pt]
\small
\noindent\textbf{Compliance Status:} COMPLIANT \\

\quad\textbf{Confidence:} 0.78\\
\textbf{Regulation:} GDPR Article 2 -- Material scope\\
\textbf{Target:} WhatsApp Terms of Service
\vspace{3pt}

\noindent\textbf{Evidence from Document.}
\begin{quote}
``WhatsApp's Privacy Policy describes our data (including message)
practices\dots and your rights in relation to the processing of information
about you.''
\end{quote}
\noindent\textbf{Explanation.} GDPR Article 2 defines the material scope of the
Regulation. The document is a Terms of Service that references a separate Privacy
Policy for data-processing practices; it neither claims that the GDPR is
inapplicable (e.g., as purely personal/household activity or competent-authority
processing) nor states that it applies where it should not. No direct conflict
with the material-scope provision is present.
\end{tcolorbox}

\begin{tcolorbox}[
  title=Example 2 -- Non-Compliant (GDPR Article 5(1)(d)),
  colback=red!3, colframe=red!70!black,
  fonttitle=\bfseries, breakable, enhanced, sharp corners, boxrule=0.6pt]
\small
\noindent\textbf{Compliance Status:} NON-COMPLIANT  \\
\quad\textbf{Confidence:} 0.62\\
\textbf{Regulation:} GDPR Article 5(1)(d) -- accuracy principle\\
\textbf{Target:} WhatsApp Terms of Service
\vspace{3pt}

\noindent\textbf{Contradiction Details.} Article 5(1)(d) requires personal data
to be accurate and kept up to date, and obliges controllers to erase or rectify
inaccurate data without delay. The document includes a broad disclaimer of
accuracy, which conflicts with this principle.

\noindent\textbf{Evidence from Document.}
\begin{quote}
``We do not warrant that any information provided by us is accurate, complete,
or useful\dots''
\end{quote}
\noindent\textbf{Explanation.} The accuracy principle is a mandatory obligation
on controllers. The Terms explicitly disclaim any commitment to accuracy, and no
other clause re-establishes a commitment to maintain or rectify accuracy; this
directly conflicts with the regulatory requirement.
\end{tcolorbox}

\begin{tcolorbox}[
  title=Example 3 -- Insufficient Information (GDPR Article 1),
  colback=orange!4, colframe=orange!80!black,
  fonttitle=\bfseries, breakable, enhanced, sharp corners, boxrule=0.6pt]
\small
\noindent\textbf{Compliance Status:} INSUFFICIENT INFORMATION\\
\textbf{Regulation:} GDPR Article 1 -- subject matter and objectives\\
\textbf{Target:} Netflix Terms of Use
\vspace{3pt}

\noindent\textbf{Evidence from Document.} None located.

\noindent\textbf{Explanation.} Article 1 concerns the protection of personal
data and the free movement of such data within the EU. The Netflix excerpt
contains no statements about how personal data is processed or protected, nor any
restriction on intra-EU personal-data movement for data-protection reasons (the
geographic restrictions it does mention relate to content licensing, not personal
data). Because the document does not speak to the obligation, compliance cannot
be determined and the system abstains rather than guessing.
\end{tcolorbox}

\begin{tcolorbox}[
  title={Example 4 -- Human Required (GDPR, general)},
  colback=blue!3,
  colframe=blue!60!black,
  fonttitle=\bfseries,
  breakable,
  enhanced,
  sharp corners,
  boxrule=0.6pt
]
\small
\noindent\textbf{Compliance Status:} HUMAN REQUIRED\\
\textbf{Regulation:} Regulation (EU) 2016/679 (general obligations)\\
\textbf{Target:} Netflix Terms of Use
\vspace{3pt}

\noindent\textbf{Evidence from Document.}
\begin{quote}
``Promotional Offers\dots We may use information such as device ID, method of
payment or your contact information to determine Offer eligibility.''\\
``\dots Customer Service may best be able to assist you by using a remote access
support tool through which we have full access to your computer. If you do not
want us to have this access, you should not consent\dots''
\end{quote}
\noindent\textbf{Explanation.} The document contains personal-data-related
statements (device ID/payment/contact information for offer eligibility; a
consent-gated remote-access support tool). None of these clearly permits an
action the GDPR forbids, and none clearly satisfies the obligation either; with
confidence below the $0.70$ threshold, the case is deferred to a human expert
rather than decided automatically.
\end{tcolorbox}

\noindent Together these examples show the full label space in action: the system
issues definite verdicts with supporting evidence where the document is
informative, and explicitly abstains or defers where it is not---the behaviour
that underlies the calibrated label distribution reported in the main text.

\section{Implementation Details}
\label{app:impl}
\textsc{ARCCS} is implemented in Python as two modules, RPEM and CCM, exposed through both
a lightweight web application and a research notebook. The RPEM parses regulatory
PDFs with the \texttt{unstructured} library (high-resolution strategy, backed by
\texttt{poppler}), converts the parsed elements to markdown, and segments the
text into sections by headers, articles, and chapters. Each section is passed to
the language model, which returns \emph{typed, schema-constrained} regulatory
objects; a quality-scoring stage then filters out incomplete or non-actionable
items below a configurable threshold. The CCM performs contradiction-based
compliance reasoning: for each requirement it retrieves the relevant document
evidence, conditions the decision on that evidence, and emits a label, a
confidence score in $[0,1]$, the supporting evidence quotes, and a justification.
The four-way decision follows a fixed priority---a detected contradiction yields
\emph{Non-Compliant}; absence of relevant information yields \emph{Insufficient
Information}; a confidence below $0.60$ yields \emph{Human Required}; otherwise
\emph{Compliant}.

The web layer is a Flask application with a browser front-end (a dark-themed
three-step wizard with drag-and-drop upload, a results table, and a per-row
detail view). Pipeline progress is streamed to the client in real time via
Server-Sent Events. Reports are stored as structured JSON and can be exported by
the user; in addition, the system maintains a per-run history (timestamped
snapshots plus a \texttt{latest} pointer per case) that supports auditing and
cross-run comparison. The application runs entirely on the user's own machine
(\texttt{python app.py}, served locally); the user supplies their own model API
key, which is read from the local environment and never bundled with the code, so
that confidential regulatory and proposal documents remain on-premise. End-to-end
latency is dominated by language-model calls and is reduced in practice by
caching pre-extracted regulation sets (e.g., the preloaded GDPR), which makes
repeated checks against the same regulation near-instant on the extraction side.

\subsection{Error Analysis}
\label{sec:erroranalysis}
The procurement benchmark has deterministic ground truth, so we can localise every
error. For the strongest backbone (\texttt{gpt-5-mini}) all $14$ residual errors
are false positives and \emph{zero} are false negatives---no genuine violation is
ever missed. Crucially, the errors are not spread across the rule set: nine of the
twelve rules are perfect ($\text{P}=\text{R}=1.0$), and all false positives
concentrate on three time-dependent rules (Table~\ref{tab:errors}). These are
exactly the constraints that require date arithmetic or reasoning about whether a
required event \emph{eventually} occurs (the $70$-day award window, the
``publication/participation must lead to an award'' rule, and a lifecycle-timing
rule). The failure mode is thus narrow and interpretable---conservative
over-flagging on temporal reasoning---rather than broad unreliability, and it
points to temporal normalisation as the most promising target for future
improvement.
 
\begin{table}[t]
\centering
\small
\setlength{\tabcolsep}{5pt}
\begin{tabular}{lcccc}
\toprule
\textbf{Rule} & \textbf{FP} & \textbf{FN} & \textbf{Prec.} & \textbf{Rec.} \\
\midrule
R06 -- award within 70 days        & 7 & 0 & 0.42 & 1.00 \\
R09 -- end-after-pub.\ lifecycle   & 5 & 0 & 0.50 & 1.00 \\
R05 -- pub./part.\ require award    & 2 & 0 & 0.83 & 1.00 \\
\midrule
Other 9 rules (each)               & 0 & 0 & 1.00 & 1.00 \\
\bottomrule
\end{tabular}
\caption{Per-rule error breakdown on the procurement benchmark
(\texttt{gpt-5-mini}; 100 cases/rule). All residual errors are false positives on
three temporal rules; no rule produces a false negative.}
\label{tab:errors}
\end{table}

\section{Model Cards}
\label{app:model-cards}
For full reproducibility, Table~\ref{tab:model-cards} lists the exact model
snapshots used in our experiments, together with the role each model plays in
the study. Unless otherwise stated, the \textsc{ARCCS} backbone is \texttt{gpt-5.2} with
reasoning effort set to \emph{medium}, and the confidence threshold for emitting
the \emph{Human Required} label is fixed at $0.60$. All models are accessed
through the OpenAI API; no model-specific fine-tuning is performed.

\begin{table}[h]
\centering
\small
\setlength{\tabcolsep}{4pt}
\begin{tabular}{ll}
\toprule
\textbf{Model} & \textbf{Snapshot} \\
\midrule
\texttt{gpt-5.2}     & \texttt{gpt-5.2-2025-12-11} \\
\texttt{gpt-5.1}     & \texttt{gpt-5.1-2025-11-13} \\
\texttt{gpt-5.2-pro} & \texttt{gpt-5.2-pro-2025-12-11} \\
\texttt{gpt-5.4}     & \texttt{gpt-5.4-2026-03-05} \\
\texttt{gpt-5-mini}  & \texttt{gpt-5-mini-2025-08-07} \\
\bottomrule
\end{tabular}
\caption{Exact model snapshots used in the experiments.}
\label{tab:model-cards}
\end{table}

\section{Full List of Procurement Rules}
\label{app:rules}
The twelve rules used in the public-procurement benchmark
(\S\ref{sec:setup-proc}) are derived directly from the statutory text of
Articles~4, 48, 56, and 73 of Directive~2014/24/EU and encode the operational
semantics of those provisions as checkable constraints. In full, the rules are:

\begin{enumerate}\setlength{\itemsep}{1pt}
    \item maximum contract-amount threshold;
    \item prohibition of duplicate publication of the same call;
    \item prohibition of award before publication;
    \item prohibition of award before participation;
    \item requirement that publication and participation eventually lead to an
    award;
    \item maximum delay of 70 days between publication and award;
    \item prohibition of contract start without prior award;
    \item prohibition of contract end before publication;
    \item prohibition of contract end immediately after publication without
    participation and award;
    \item requirement that any started contract must eventually end;
    \item requirement that any ended contract must previously have started;
    \item requirement that any terminated contract must previously have been
    awarded.
\end{enumerate}

\noindent These rules span four compliance dimensions: monetary constraints
(rule~1), duplicate-publication restrictions (rule~2),
temporal/logical consistency of award procedures (rules~3--6), and lifecycle
consistency for contract execution and termination (rules~7--12).

\section{Terms-of-Use Documents}
\label{app:sizes}

We use the GDPR as the regulatory input and evaluate three widely used
terms-of-use documents that differ markedly in length and specificity:
\emph{WhatsApp Terms of
Service},\footnote{\url{https://www.whatsapp.com/legal/terms-of-service/revisions/20210104}}
\emph{Netflix Terms of
Use},\footnote{\url{https://help.netflix.com/legal/termsofuse?locale=en-GB}}
and \emph{ChatGPT Terms of
Use}.\footnote{\url{https://openai.com/en-GB/policies/row-terms-of-use}}

Table~\ref{tab:sizes} reports basic size statistics for these documents. The
three documents differ substantially in length, allowing us to examine how the
amount of available textual evidence affects the resulting label distribution.
In particular, the much longer WhatsApp document provides more evidence for the
model to ground its decisions, leading to more determinate compliance outcomes
than those obtained for the shorter Netflix and ChatGPT documents.
\begin{table}[h]
\centering
\small
\setlength{\tabcolsep}{4pt}
\begin{tabular}{lrrrrr}
\toprule
\textbf{Document} & \textbf{Chars} & \textbf{Words} & \textbf{Sent.} & \textbf{Paras} & \textbf{Lines} \\
\midrule
WhatsApp & 323{,}682 & 51{,}698 & 1{,}862 & 1{,}382 & 3{,}723 \\
Netflix  & 15{,}777  & 2{,}538  & 128     & 39      & 158 \\
ChatGPT  & 20{,}404  & 3{,}324  & 171     & 100     & 197 \\
\bottomrule
\end{tabular}
\caption{Terms-of-use document sizes used in our experiments.}
\label{tab:sizes}
\end{table}

\section{Public-Procurement Benchmark Details}
\label{app:proc-benchmark}

The controlled public-procurement benchmark is grounded in Directive~2014/24/EU
(CELEX:~32014L0024), using twelve rules derived from Articles~4, 48, 56, and 73.
The rules operationalise four compliance dimensions: monetary constraints,
duplicate-publication restrictions, temporal/logical consistency of award
procedures, and lifecycle consistency for contract execution and termination.

Each benchmark case follows the Tenders Electronic Daily (TED) event-log schema
and is represented as a temporally ordered trace over the events
\texttt{PUBLICATION}, \texttt{PARTICIPATION}, \texttt{AWARD},
\texttt{CONTRACT-START}, and \texttt{CONTRACT-END}. Traces include
procurement-relevant attributes such as contract amount, procedure type,
electronic-handling and framework-agreement indicators, country, NUTS code, CPV
division and code, and trace length.

Ground-truth labels are assigned through deterministic checks over the event
sequence and attributes. For example, an \texttt{AWARD} event preceding
\texttt{PUBLICATION} violates the ordering rule, while a trace containing
\texttt{CONTRACT-START} without a corresponding \texttt{CONTRACT-END} violates
the lifecycle-completion rule. This construction ensures that evaluation labels
are independent of \textsc{ARCCS} outputs and of any LLM-as-a-judge procedure.

latex \section{Compliance Label Distribution} \label{app:label-distribution}  Table~\ref{tab:label-dist} reports the distribution of compliance labels over the 100 GDPR-derived requirements for each evaluated terms-of-use document. The distribution supports the interpretation in Section~\ref{sec:results}: abstention is the dominant outcome because public terms-of-use documents typically do not disclose the internal governance evidence required to assess many GDPR obligations. In this setting, \emph{Insufficient Information} is not a failure mode by itself; it is the expected outcome when the target document does not contain the evidence needed for a faithful decision.  The distribution also shows that \textsc{ARCCS} does not abstain uniformly. The longer WhatsApp document provides more retrievable evidence and therefore leads to more determinate outcomes and more cases routed to \emph{Human Required}. By contrast, shorter or less informative documents leave fewer requirements sufficiently evidenced. This pattern is consistent with the intended behaviour of the label scheme: definite \emph{Compliant} and \emph{Non-Compliant} decisions are reserved for cases supported by retrieved evidence, while uncertain or underspecified cases are not forced into artificial verdicts.  \begin{table}[t] \centering \small \setlength{\tabcolsep}{4pt} \begin{tabular}{lcccc} \toprule \textbf{Document} & \textbf{Comp.} & \textbf{Non-C.} & \textbf{Insuf.} & \textbf{Human} \\ \midrule WhatsApp & 2 & 2 & 89 & 7 \\ Netflix  & 0 & 0 & 99 & 1 \\ ChatGPT  & 0 & 4 & 94 & 2 \\ \bottomrule \end{tabular} \caption{Compliance-label distribution over the 100 GDPR-derived requirements for each evaluated terms-of-use document.} \label{tab:label-dist} \end{table}  The same distribution should be read together with the qualitative conflict examples discussed in the main text. The WhatsApp and ChatGPT cases illustrate that calibrated abstention does not make the system passive: when the document contains clauses that conflict with a requirement, \textsc{ARCCS} can still surface a candidate violation with supporting evidence and explanation. The resulting flags are intended for expert review rather than automatic legal determination.

\section{Intended Users and Deployment}
\label{app:users}

\textsc{ARCCS} is designed for practitioners who need to assess documents against regulatory requirements but are not necessarily machine-learning engineers. Its intended users include in-house legal and compliance teams, data protection officers, public-sector procurement officers, and smaller organisations that lack dedicated ML infrastructure. For such users, practical usability is as important as technical capability: the system must be easy to install, easy to operate, and able to support compliance review without requiring users to write or adapt code.

To address this need, \textsc{ARCCS} is released as a free and fully open-source application that can be installed and run locally as a lightweight web interface. Once deployed, users interact with the system through the browser: they upload a regulation, provide a target document, launch the analysis, and inspect the resulting compliance report through the interface. The application therefore exposes the full compliance-checking pipeline as an interactive workflow, while hiding the underlying orchestration, prompting, retrieval, and report-generation steps from the end user.

This design makes \textsc{ARCCS} suitable for users with different levels of technical expertise. Legal and compliance practitioners can use the interface to obtain structured, auditable reports, while technical users can inspect, adapt, or extend the open-source codebase for their own deployment requirements. The system stores run history and generated reports locally, supports repeated analyses over different documents, and allows users to review each decision together with its associated requirement, confidence score, evidence, and explanation.

To our knowledge, \textsc{ARCCS} is the first freely available, fully open-source system offering end-to-end, regulation-agnostic compliance checking in this form. By combining a locally runnable application, an interactive interface, and structured compliance reports, it lowers the barrier to automated regulatory review for the non-specialist users who are most likely to benefit from such tools.

\end{document}